\documentclass[11pt,a4paper]{article}
\usepackage[hyperref]{eacl2021}
\usepackage{times}
\usepackage{latexsym}

\usepackage{graphicx}  
\usepackage{microtype}

\aclfinalcopy 
\def\aclpaperid{698} 

\title{Few Shot Dialogue State Tracking using Meta-learning}

\author{Saket Dingliwal\textsuperscript{1,2} $\,\,\,$Shuyang Gao\textsuperscript{1} $\,\,\,$   Sanchit Agarwal\textsuperscript{1}  $\,\,\,$  Chien-Wei Lin\textsuperscript{1} \\
 \bf  Tagyoung Chung\textsuperscript{1} $\,\,\,$ 
Dilek Hakkani-T{\"u}r\textsuperscript{1}\\
\textsuperscript{\rm 1}Amazon Alexa AI $\;\;\,\,$
\textsuperscript{\rm 2}Carnegie Mellon University\\ 

\texttt{{\{skdin,shuyag,agsanchi,chienwel,tagyoung,hakkanit\}}@amazon.com}

}

\date{}

\usepackage[ruled,linesnumbered ]{algorithm2e}
\SetKwInOut{Parameter}{Parameters}

\def\train{\textit{train }}
\def\target{\textit{target }}
\usepackage[inline]{enumitem}

\begin{document}
\maketitle
\begin{abstract}
    Dialogue State Tracking (DST) forms a core component of automated chatbot based systems designed for specific goals like hotel, taxi reservation, tourist information etc. With the increasing need to deploy such systems in new domains, solving the problem of zero/few-shot DST has become necessary. There has been a rising trend for learning to transfer knowledge from resource-rich domains to unknown domains with minimal need for additional data. In this work, we explore the merits of meta-learning algorithms for this transfer and hence, propose a meta-learner D-REPTILE specific to the DST problem. With extensive experimentation, we provide clear evidence of benefits over conventional approaches across different domains, methods, base models and datasets with significant (5-25\%) improvement over the baseline in low-data setting. Our proposed meta-learner is agnostic of the underlying model and hence any existing state-of-the-art DST system can improve its performance on unknown domains using our training strategy.
    
\end{abstract}







\section{Introduction}
Task-Oriented Dialogue (TOD) systems are automated conversational agents built for a specific goal (for example hotel reservation). Many businesses from wide-variety of \textit{domains} (like hotel, restaurant, car-rental, payments etc) have adopted these systems to cut down their cost on customer support services. Almost all such systems have a Dialogue State Tracking (DST) module which keeps track of values for some predefined domain-specific slots (example hotel-name, restaurant-rating etc) after every turn of utterances from user and system. These values are then used by Natural Language Generator (NLG) to generate system responses and fulfill the user goals. 

Many of the recent works \cite{wu2019transferable, zhang2019find, goel2019hyst, heck2020trippy} have proposed various neural models that achieve good performance for the task but are data hungry in general. Therefore, adapting to a new unknown domain (\target domain) requires large amounts of domain-specific annotations limiting their use. However, given a wide range of practical applications, there has been a recent interest in data-efficient approaches.  \citet{lee2019sumbt}, \citet{gao2020machine} used transformer \cite{vaswani2017attention} based models which significantly reduce data dependence. Further, \citet{gao2020machine} model the problem as machine reading comprehension task and benefit from its readily available external datasets and methods.  \citet{wu2019transferable} were first to propose transferring  knowledge from one domain to another. Since, many domains like restaurant and hotel share a lot of common slots like name, area, rating, etc and hence such a transfer proved to be effective for a low-resource domain. More recently, \citet{campagna2020zero} aimed at zero-shot DST using synthetic data generation for the \target domain imitating data from other domains. 

Recent meta-learning methods like MAML \cite{finn2017model}, REPTILE \cite{nichol2018first} have proven to be very successful in efficient and fast adaptations to new tasks with very few labelled samples. These methods specifically aim at the setting where there are many similar tasks but very small amount of data for each task. Agnostic of the underlying model, these meta-learning algorithms spit out initialization for its parameters which when fine-tuned using low-resource \target task achieves good performance. Following their widespread success in few-shot image classification, there has been a lot of recent work on their merit in natural language processing tasks. \citet{huang2018natural, gu2018meta, sennrich2019revisiting, bansal2019learning, dou2019investigating, yan2020multi}  attempt at using meta-learning for efficient transfer of knowledge from high-resource tasks to a low-resource task. Further, some of the more recent works \cite{dai2020learning, qian2019domain} have shown meta-learners can be used for system response generation in TOD systems which is generally downstream task for our DST task.

To the best of our knowledge, ours is the first work exploring meta-learning algorithms for the DST problem. While prior work focused on training their models with a mixture of data from other available domains (\train domains) followed by fine-tuning with data from \target domain, we identify that this method of transferring knowledge between domains is inefficient, particularly in very low-data setting with just $0, 1, 2, 4$ annotated examples from \target domain. We, on the other hand, use \train domains to meta-learn the parameters of the model used to initialize the fine-tuning process. 
We hypothesize that though different domains share many common slots, they can have different complexities. For some of the domains, it might be easier to train the model using very few examples while others may require large number of gradient steps (based on their different data complexity and training curves with 1\%, 5\%, 10\% data in \citet{gao2020machine}). 
Meta-learning takes into account that this gradient information and share it across domains. Rather than looking for an initialization that try to simultaneously minimizes joint loss over all the domains, it looks for a point from which the optimum parameters of individual domains are reachable in few ($<5$) gradient steps (and hence very few training examples for these steps). Then the hope is that the \target domain is similar to at least one of the \train domains (for example hotel \& restaurant or taxi \& train)  and hence the learned initialization will achieve efficient fine-tuning with very few examples for the \target domain as well. 
This direction of limited data is motivated by practical applicability, where it might be possible for any developer to manually annotate $4$-$8$ examples before deploying the chatbot for a new domain.

We highlight the main contributions of our work below
\begin{enumerate*}[label=(\roman*)]
    \item We are the first to explore and reason about the benefits of meta-learning algorithms for DST problem
    \item We propose a meta-learner D-REPTILE that is agnostic to underlying model and hence has the ability to improve the state of the art performance in zero/few-shot DST for new domains. 
    \item With extensive experimentation, we provide evidence of the benefit of our approach over conventional methods. We achieve a significant 5-25\% improvement over the baseline in few-shot DST that is consistent across different \target domains, methods, base models and datasets.
\end{enumerate*}

\section{Background}
\label{sec:background}
\subsection{Dialogue State Tracking}
DST refers to keeping track of the \textit{state} of the dialogue at every turn. \textit{State} of dialogue can be defined as $\langle \textit{slot\_name,slot\_value}\rangle$ pairs that represents, given a domain-specific slot, the value that the user provides or system-provided value that the user accepts. Further, many domains have a pre-defined \textit{ontology} that specify the set of values each slot can take. 
Note that the number of values in ontology varies a lot with slots. Some slots like hotel-stars might have just five different values (called \textit{categorical} slots), while those like hotel-name have hundreds of possible values (called \textit{extractive} slots). It might be possible that a slot has never been discussed in the dialogue sequence and in that case, model has to predict a \textit{None} value for that particular slot.

Various models have been proposed for the above task, but particularly relevant to this work is transformer-based model \textit{STARC} by \citet{gao2020machine}. For each slot, they form a question (like what is the name of the hotel for hotel-name slot) and then at each turn append the tokens from dialogue utterance and the question separated by [SEP] token. They then pass these sequence of tokens through a transformer to form \textit{token embeddings}. 
For the \textit{extractive} slots, they use \textit{token embeddings} to mark the span (start and end position) of the answer value in the dialogue itself (called \textit{extractive-model}). For the \textit{categorical} slots with less number of possible values, \textit{categorical-model} append embeddings of each possible value to the \textit{token embeddings} and then use a classifier with softmax layer to predict the correct option. 

\subsection{Meta-Learning}
With advances in model-agnostic meta-learning framework by \citet{finn2017model, nichol2018first}, the few-shot problems have been revolutionized. These frameworks define a underlying task-distribution from which both train ($\tau$) and target tasks ($\tau^{'}$) are sampled. For each task $\tau$, we are given very few labelled data-points $\mathcal{D}^{train}_{\tau}$ and a loss function $\mathcal{L}_{\tau}$. Now, given a new data point $\mathcal{D}^{test}_{\tau^{'}}$ from target task $\tau^{'}$, the goal is to learn parameters $\theta_{\mathcal{M}}$ of any model $\mathcal{M}$ such that $\mathcal{L}_{\tau^{'}}(\mathcal{D}^{test}_{\tau^{'}}; \theta_{\mathcal{M}})$ is minimized. This is achieved by $k$-steps of gradient descent using $\mathcal{D}^{train}_{\tau^{'}}$ with learning rate $\alpha$. More formally, $\theta_{\mathcal{M}} = \textit{SGD}(\mathcal{D}^{train}_{\tau^{'}}, \mathcal{L}_{\tau^{'}}, \theta_{\mathcal{M}}^{INIT}; k, \alpha)$ where $\textit{SGD}(\mathcal{D}, \mathcal{L}, \theta^{INIT}; k, \alpha)$ gives $\theta^{(k)}$ such that 
\begin{equation} \label{eq:sgd}
  \theta^{(t)} = \theta^{(t-1)} - \alpha \nabla_{\theta}(\mathcal{L}(\mathcal{D}; \theta)), \theta^{(0)} = \theta^{INIT}  
\end{equation}
Therefore, the goal now is to find a good initialization $\theta_{\mathcal{M}}^{INIT}$ for the gradient descent using the data from train tasks $\tau$. This is achieved by minimizing the empirical loss as
\begin{equation} 
    \theta^{(k)} = \textit{SGD}(\mathcal{D}^{train}_{\tau}, \mathcal{L}_{\tau}, \theta; k, \alpha) \\
\end{equation}
\begin{equation}
    \label{eq:obj}
    \theta_{\mathcal{M}}^{INIT} = arg\,\min_\theta \sum_{\tau}  \mathcal{L_{\tau}}(\mathcal{D}^{train}_{\tau}; \theta^{(k)})
\end{equation}

Note that the above optimization is complex and involve second-order derivatives. For computational benefits, \citet{nichol2018first} proposed REPTILE and showed that these terms can be ignored without effecting the performance of the meta-learning algorithm. We refer the reader to their work for more details.

\section{Methodology}
\label{sec:meth}
In this work, we propose D-REPTILE, a meta-learning algorithm specific to DST task. Following what \citet{qian2019domain} did for dialogue generation problem, we treat different domains as tasks for the meta-learning algorithm. Let $\mathrm{D} = \{d_1, d_2, \dots d_n\}$ (eg. $\{restaurant, taxi, payment, \dots\})$ be the set of \train domains for which we have annotated data available. Let $p_D(.)$ define a probability distribution over these domains. Let $\mathcal{D}_{d_1}, \mathcal{D}_{d_2} \dots \mathcal{D}_{d_n}$ be the training data from each of these domains. Let $\mathcal{M}$ be any DST model with parameters $\theta_{\mathcal{M}}$. Let $m$ be the task-batch size (number of domains in a batch in our case), $\alpha, \beta$ be the inner and outer learning rate respectively, $k$ be the number of gradient steps. Let $\textit{SGD}(.)$ be the function as defined in equation \ref{eq:sgd}. 
Borrowing the meta-learning theory regarding optimizing the objective equation \ref{eq:obj} from \citet{nichol2018first}, we define the algorithm D-REPTILE in Algorithm \ref{alg:dreptile}. The update rule for initialization (as defined in step 8) is same as that of REPTILE. We chose REPTILE over other meta-learning algorithms because of its simplicity and computational advantages. Nonetheless, its straight-forward to switch any other initialization based meta-learner by changing meta-update step. The novelty of our learner lies in its definition of the meta-learning tasks that represent different domains of DST problem. This algorithm aims to find $\theta_{\mathcal{M}}^{INIT}$, which we use to initialize the model for the fine-tuning stage with the \target domain.

\begin{algorithm}
\SetAlgoLined
\SetKwInput{KwData}{Input}
\KwData{$\mathcal{D}_{d_1}, \mathcal{D}_{d_2} \dots \mathcal{D}_{d_n}$}
\Parameter{$\mathcal{M}$, $\mathcal{L}$, $p_D(.)$, $\alpha$, $\beta$, $k$, m}
\KwResult{$\theta_{\mathcal{M}}^{INIT}$}
Initialize $\theta_{\mathcal{M}}$ randomly \\
\For{iteration $i = 1,2,\dots $}{
    sample m domains $\mathrm{D_i}$ using $p_D(.)$\\
    \For{domain $d_j \in \mathrm{D}_i$}{
        sample data points $\mathcal{D}_{ij}$ from $\mathcal{D}_{d_j}$\\
        $\theta_{\mathcal{M}}^{d_j} \leftarrow \textit{SGD}(\mathcal{D}_{ij}, \mathcal{L}, \theta_{\mathcal{M}}; k, \alpha)$ 
    }
    $\theta_{\mathcal{M}} \leftarrow \theta_{\mathcal{M}} + \beta \frac{1}{m} \sum_{j=1}^{m} (\theta_{\mathcal{M}}^{d_j} - \theta_{\mathcal{M}})$
}
return $\theta_{\mathcal{M}}$; \\
 \caption{D-REPTILE: Meta Learner for DST}
\label{alg:dreptile}

\end{algorithm}

We argue that the meta-learned initialization are better suited for fine-tuning than conventional methods. In the hope that joint optimal parameters for \train domains lie close to individual domains, \citet{wu2019transferable} initialize the fine-tuning stage of the \target domain from the joint minimum of the loss from data from all the \train domains (called Naive pre-training before Fine-Tuning or NFT here). More formally, they chose the following initialization 
\begin{equation} \label{eq:tl}
    \theta_{\mathcal{M}}^{INIT} = arg \min_\theta \sum_{j=1}^{n} \mathcal{L}(\mathcal{D}_{d_j};\theta)
\end{equation}
Such an initialization tries to simultaneously minimize the loss for all the domains which might be useful if the goal was to perform well on test data coming from mixture of these domains. However, here our goal is to perform well on a single unknown \target domain and no direct relation between this initialization and the optimal parameters for the \target domain can be seen. Further, as the number of \train domains increases or training data for each  domain decreases, the joint optimum can be very far-off from the individual domain-optimum parameters. Therefore, these methods perform particularly bad. We show empirical evidence for this hypothesis in Section~\ref{sec:exp}. On the other hand, if we optimize equation \ref{eq:obj}, we will reach a point in the parameter space from where all the domain-optimum parameters are reachable in k-gradient descent steps. Therefore, we can hope to reach the optimum parameters for the \target domain as well efficiently. This hope is  much larger for DST problem specifically because of similarities in different related domains (specifically related slots as shown in Section~\ref{sec:ablation}).

Let us consider the following example, let restaurant and taxi be two of the \train domains. Optimizing equation \ref{eq:obj}, we might reach a point which is closer to optimum parameters of restaurant domain than taxi domain if we have have smaller gradient values for restaurant data but large for taxi. However notably, both the optimum-parameters are reachable in k-gradient steps. Now if \target domain is hotel (similar to restaurant domain with common slots like rating, name, etc), we will already be close to its optimum parameters. Also if the \target domain is bus (similar to taxi domain with common slots like time, place, etc), we will have larger gradients in fine-tuning stage and thus will reach the optimum parameters for bus as well. This might not have been possible with equation \ref{eq:tl} as the optimum parameters for joint of restaurant and taxi data might be very far from both the individual \train domains and will also have no specific gradient properties for faster adaptation for any of hotel or bus \target domains.

\section{Experiments}
\label{sec:exp}
\subsection{Datasets}
We used two different DST datasets for our experiments. (i) MultiWoz 2.0 \cite{budzianowski2018multiwoz}, 2.1 \cite{eric2019multiwoz} (ii) DSTC8 \cite{rastogi2019towards}. The former is manually annotated complex dataset with mostly 5 different domains, 8438 dialogues while the latter is relatively simple synthetically generated dataset with 26 domains and 16142 dialogues. Both the datasets contains dialogues spanning multiple domains. Following the setting from \citet{wu2019transferable}, for extracting data of a particular domain from the dataset, we consider all the dialogues in which that domain is present and ignore slots from other domains both in train and test set. Further, as shown by \citet{gao2020machine},  we use external datasets from Machine Reading for Question Answering (MRQA)
2019 shared task \cite{fisch2019mrqa}, DREAM \cite{sun2019dream}, RACE \cite{lai2017race} to pre-train our transformer in our experiments and label it with suffix '-RC' to distinguish it from '-base' model.
\subsection{Evaluation Metric}
Based on the objective in DST, there is a well established metric Joint Goal Accuracy (JGA). JGA is the fraction of total turns across all dialogues for which the predicted and ground truth dialogue state matches for all slots. Following \citet{wu2019transferable}, for testing for a single \target domain in a multi-domain dialogue, we only consider slots from that domain in metric computation. Note that in some of our experiments (where explicitly mentioned), we further restrict the slots to only extractive or only categorical slots. 
Also, as it happens most of the times, whenever a slot is not mentioned in any turn, the ground truth value for that slot is \textit{None}. For analysis, we further use the metric Active Slot Accuracy which is the fraction of predicted values of a particular slot that were correct whenever the ground truth value was not \textit{none}.

\subsection{Experimental Setting}
For all our experiments, both D-REPTILE and baseline (NFT (Sec. \ref{sec:meth})) uses STARC  (Sec. \ref{sec:background}) as base model $\mathcal{M}$. This ensures that all the gains achieved in our experiments are only due to meta-learning. In our implementation \footnote{\href{https://github.com/saketdingliwal/Few-Shot-DST}{https://github.com/saketdingliwal/Few-Shot-DST}}, we use pre-trained word embeddings Roberta-Large \cite{liu2019roberta}, Adam optimizer \cite{kingma2014adam} for gradient updates in both inner and outer loop, $\alpha = 5e^{-5}, \, \beta = 1, \, m=4, \, k=5, \, p_D(i) \propto |\mathcal{D}_{d_i}|$ (chosen using dev-set experiments as explained in Section \ref{sec:ablation}). 
As shown recently \cite{mosbach2020stability}, the fine-tuning of transformer based model is unstable, therefore, we run fine-tuning multiple times and report the mean and the standard deviation of the performance. Also, the performance varies with the choice of training data from \target domain used for fine-tuning. However, for our experiments, we chose these dialogues based on number of active slots (not \textit{None}) and use the same dialogue for both D-REPTILE and baseline. Since, we use very little data (0, 1, or 2 examples) from \target domain, we obviously would like to have  dialogues that at-least have all the slots being discussed in the utterances. In practical scenarios, where a developer might be creating 1 or 2 examples for a new domain, it is always possible to include all the slots in the dialogue utterances.


\subsection{Results}
\begin{figure*}
    \centering
    \includegraphics[width=  \linewidth]{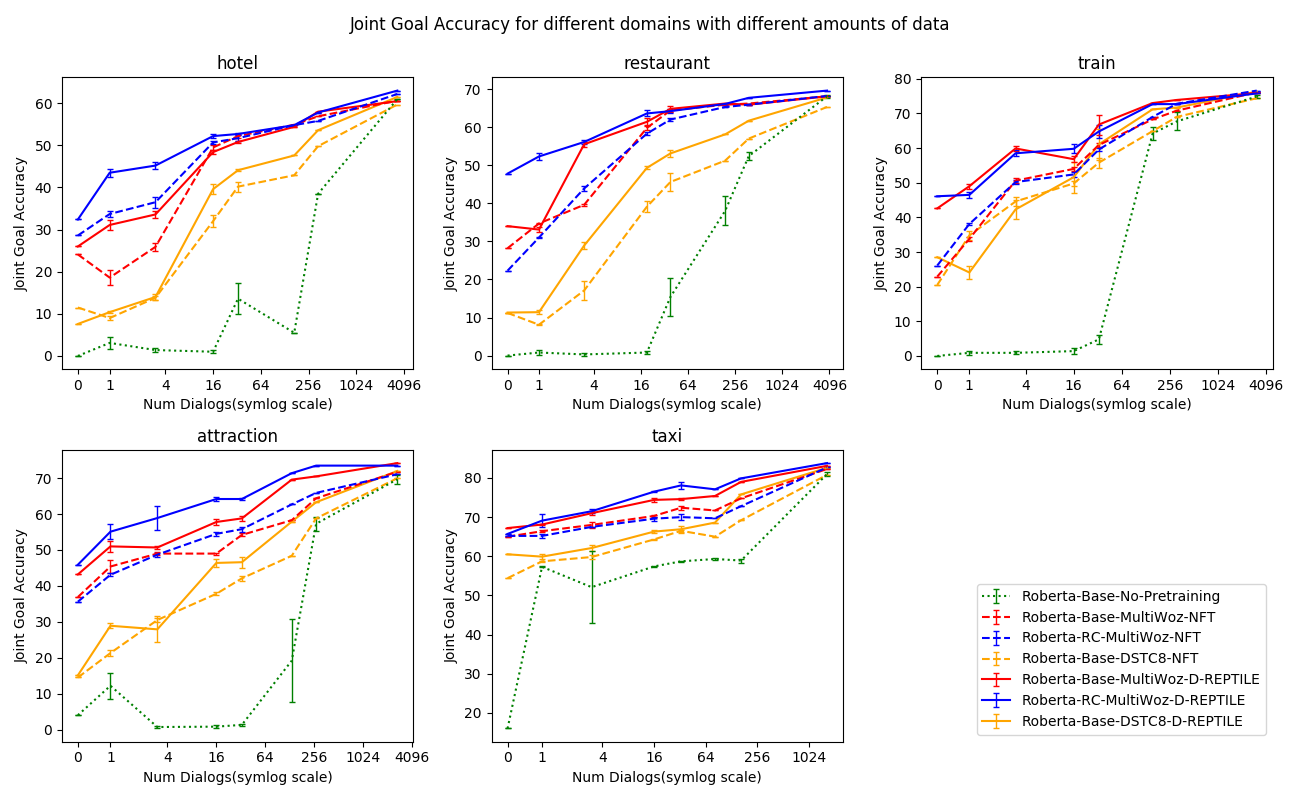}
    \caption{Performance of D-REPTILE vs NFT for different MultiWoz domains with three different models. 
    }
    \label{fig:multiwoz}
\end{figure*}


In our experiments, we are able to achieve significant improvement over the baseline method under low-data setting ($<$ 32 dialogues). 
Note that the choice of low-data setting is guided by the practical applications of the method. It also validates our hypothesis that the initialization chosen by meta-learning is closer to optimal parameters of the \target domain in terms of gradient steps and therefore perform better when there is very less data. However,
as fine-tuning data is increased to 1000s of dialogues, any random initialization is also able to reach the optimal parameters for \target domain. We observe the benefits of D-REPTILE in limited data consistently across different domains, datasets and models as explained one-by-one below

\textbf{Across domains} - We used all different domains of MultiWoz 2.0 data as \target domain in 5 plots in Figure \ref{fig:multiwoz}. We pre-train D-REPTILE (solid) and NFT (dashed) versions of different models (represented by different colors). For the models represented by red and blue colors, we used all domains other than \target domain as our \train domains. For example, for the first plot, hotel domain is our \target domain, while restaurant, train, attraction and taxi are our \train domains. The red corresponds to  starting with Roberta-Base embeddings, while the blue represent Roberta-RC which is pre-trained Roberta-Base with reading comprehension datasets \cite{gao2020machine}. The green dotted line represent model without any pretraining. It is clearly very bad and unstable. This shows importance of using other domains for few-shot experiments. We fine-tune all our models using different amount of training data of \target domain (x-axis).  In each one of our models, the solid lines (D-REPTILE) lies strictly above the dashed lines (NFT) in JGA metric. The gains obtained are as high as 47.8\% (D-REPTILE) vs 22.3\% (NFT) for restaurant domain with 1 dialogue which is more than 100\% improvement at no annotation cost at all.

\begin{figure}
    \centering
    \includegraphics[width=\linewidth]{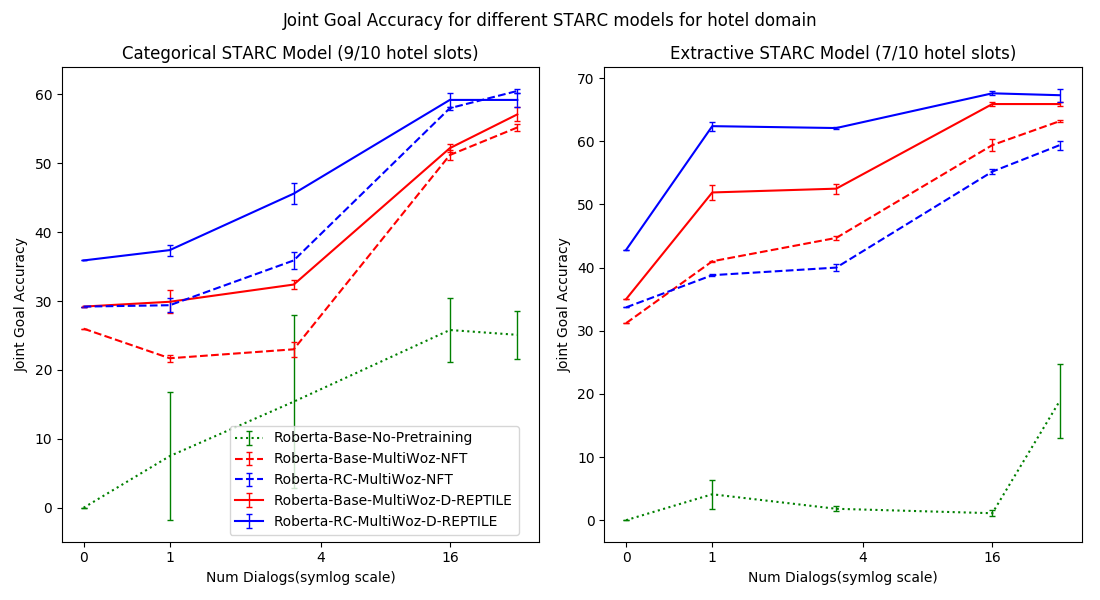}
    \caption{Performance of D-REPTILE vs NFT for different DST models for different slots in hotel domain.
    }
    \label{fig:cat_non_cat}
\end{figure}

\textbf{Across models} - Not only the results are consistent across different base models for transformer as shown in Figure \ref{fig:multiwoz} but also across different DST methods. As done in \citet{gao2020machine}, we train separate categorical and extractive models for hotel domain (using categorical and extractive data respectively from \train domains) (which we have combined to  plot Figure \ref{fig:multiwoz}). If we consider these two fairly different models separately, we achieve similar trends in each individually as plotted in Figure \ref{fig:cat_non_cat}. Note that JGA metric is computed here with restricted slots based on the type of the model. The gains are larger for the extractive model possibly because marking span in original dialogue can be considered slightly harder task than choosing among limited number of choices. 

\begin{figure}
    \centering
    \includegraphics[width=0.8\linewidth]{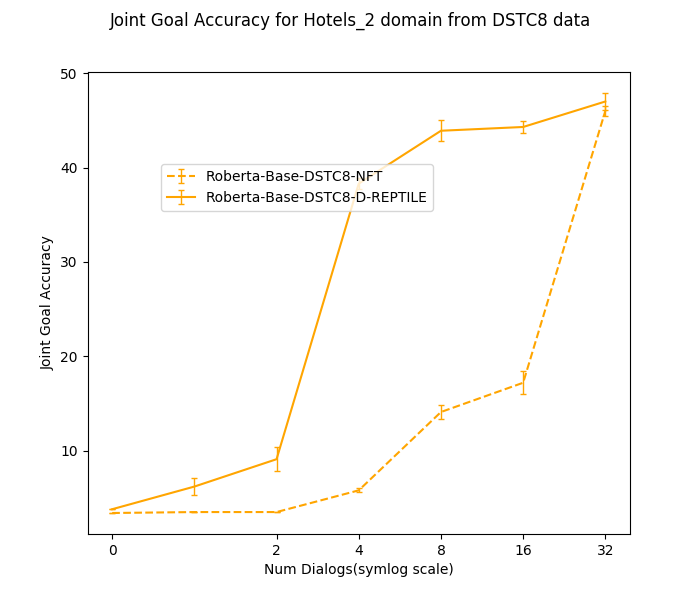}
    \caption{Performance of D-REPTILE vs NFT for Hotels\_2 domain in DSTC8 data as \target domain. 
    }
    \label{fig:cat_dstc8}
\end{figure}

\textbf{Across datasets} - To show that the merits of D-REPTILE are not limited to MultiWoz data, we tested with domains from DSTC8 dataset as both \train and \target domain. In Figure \ref{fig:multiwoz}, the orange lines represent model pre-trained using all the domains in DSTC8 as \train domains while \target domain is from MultiWoz. As expected, the performance of these models fall below red and blue lines (models pre-trained with MultiWoz \train domains) but above green (no pre-training) as training and testing datasets are different.  However, the solid orange line (D-REPTILE) lies above dashed line (NFT). In another set of experiments, we used \target domain from DSTC8 and compiled the results in Figure \ref{fig:cat_dstc8}. Except for Hotels\_1, Hotels\_2 and Hotels\_3, all other domains from DSTC8 are used as \train domains while Hotels\_2 is kept as \target domain. We see that the benefits of meta-learning are much larger for DSTC8 dataset than MultiWoz.  For example, with 8 dialogues for fine-tuning, D-REPTILE achieves JGA of 43.9\% while NFT is only able to get 14.1\%. This can be attributed to increase in number of different training tasks (23 domains were used as \train domains for DSTC8 as opposed to 4 for MultiWoz experiments).

Surprisingly, the meta-learned initializations not only adapt faster but are also better to start with. We see an improvement in zero-shot performance as well. In addition to comparison with the NFT baseline, we also show improvement over existing models on MultiWoz 2.0 dataset in Table \ref{tab:zero}. Also note that D-REPTILE is model-agnostic and therefore has the capability to improve the JGA for any underlying model for a new unknown domain. 


\begin{table}[]
\resizebox{\linewidth}{!}{%
\begin{tabular}{l c c c c c}
\hline
Joint Goal Accuracy &
  \multicolumn{1}{l}{\textbf{Hotel}} &
  \multicolumn{1}{l}{\textbf{Restaurant}} &
  \multicolumn{1}{l}{\textbf{Taxi}} &
  \multicolumn{1}{l}{\textbf{Attraction}} &
  \multicolumn{1}{l}{\textbf{Train}} \\ \hline
TRADE \cite{wu2019transferable}           & 13.7          & 11.52         & 60.58         & 19.87         & 22.37         \\ \hline
STARC \cite{gao2020machine}          & 28.6          & 28.2          & 65            & 36.9          & 26.1          \\
STARC + D-REPTILE & \textbf{32.4} & \textbf{47.8} & \textbf{67.2} & \textbf{45.9} & \textbf{46.1} \\ \hline
\end{tabular}
}
\caption{\label{tab:zero} Zero-shot performance on MultiWoz 2.0 dataset. Domains like restaurant and train witness a significant boost in performance over baselines}
\end{table}
\section{Ablation Studies}
\label{sec:ablation}

To validate our various theoretical hypothesis, search for hyper-parameters, clearly identify and reason about the situations where using meta-learning helps DST, we perform additional analysis as written in subsections below.


\subsection{Slot-wise Analysis}
\begin{figure*}
    \centering
  \includegraphics[width=\linewidth]{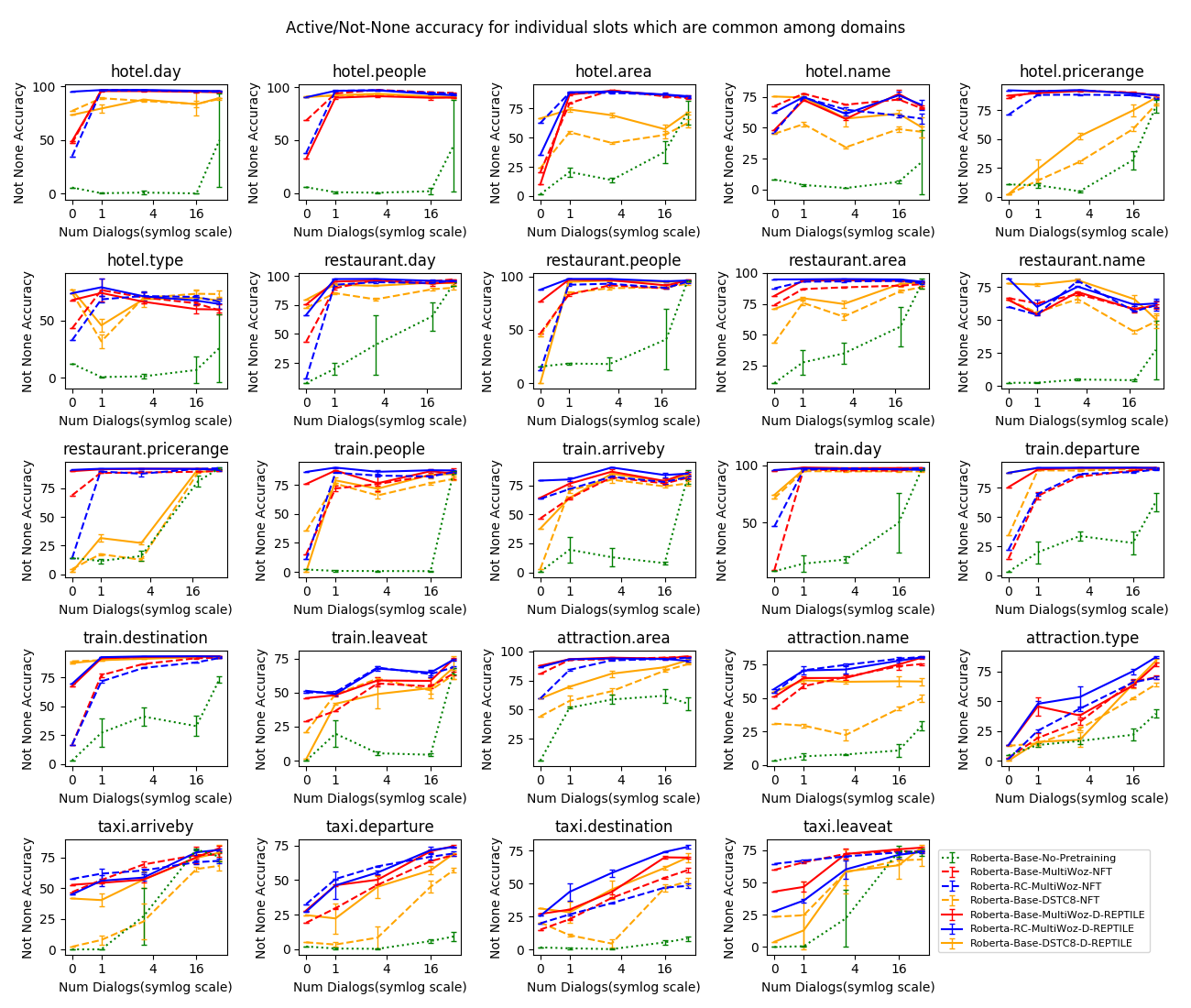}
  \caption{Active Slot accuracy for slots common between different domains}
  \label{fig:slot_wise_common}
\end{figure*}

\begin{figure*}
    \centering
  \includegraphics[width=\linewidth]{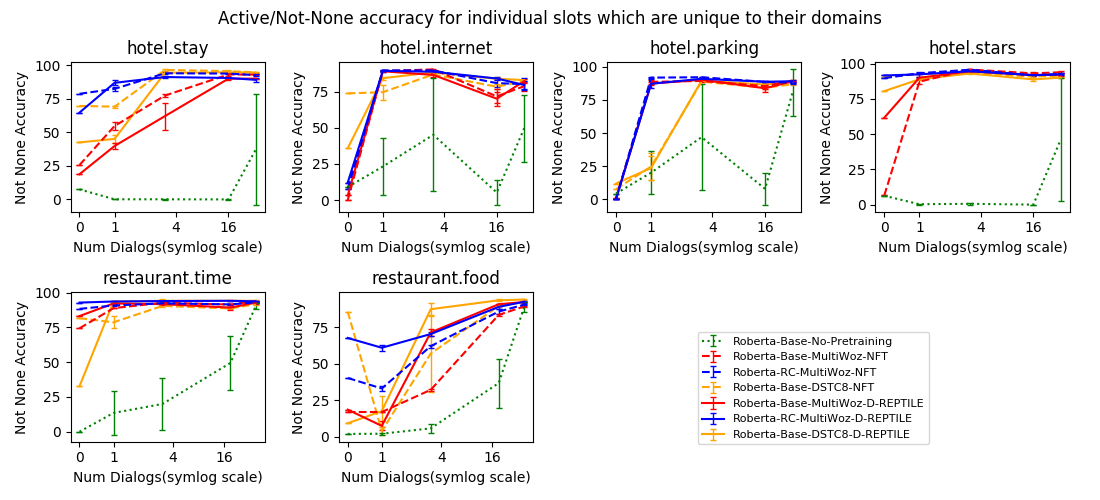}
  \caption{Active Slot accuracy for slots unique to specific \target domain}
  \label{fig:slot_wise_unique}
\end{figure*}

To exactly pin-point the advantage of D-REPTILE, we do a slot-wise analysis of our models in Figure \ref{fig:slot_wise_common} and \ref{fig:slot_wise_unique}. Note that slots are  defined as $\textit{domain\_name.slot\_name}$. For example, \textit{hotel.day} represents performance of the models in predicting the values for \textit{day} slot where the \target domain was \textit{hotel}. Overall performance or JGA in plot 1 of Figure \ref{fig:multiwoz} is combination of all the \textit{hotel} slots like \textit{day, people, area, etc}. Figure \ref{fig:slot_wise_common} shows the slots which are common among different domains while Figure \ref{fig:slot_wise_unique} compare the performance for slots that are unique to a \target domain. We can see that for the common slots, the solid lines (D-REPTILE) mostly lie higher than the dashed (NFT) counterparts. However, nothing can be said in particular about slots in Figure \ref{fig:slot_wise_unique}. This behaviour is expected as unique slots particular to a \target domain have little to gain from the different slots present in \train domains (which were used for pre-training). This is evident from the fact that slots like \textit{hotel.internet}, \textit{hotel.parking} have zero-shot active accuracy close to zero for all kinds of pretraining strategies (Figure \ref{fig:slot_wise_unique}). 
However, wherever slots between different domains are similar, the pretraining have much larger influence. In that case, the merit of learning generalizable initialization from D-REPTILE than NFT is much more clearly evident (Figure \ref{fig:slot_wise_common}).
\begin{figure}
    \centering
  \includegraphics[width=0.85\linewidth]{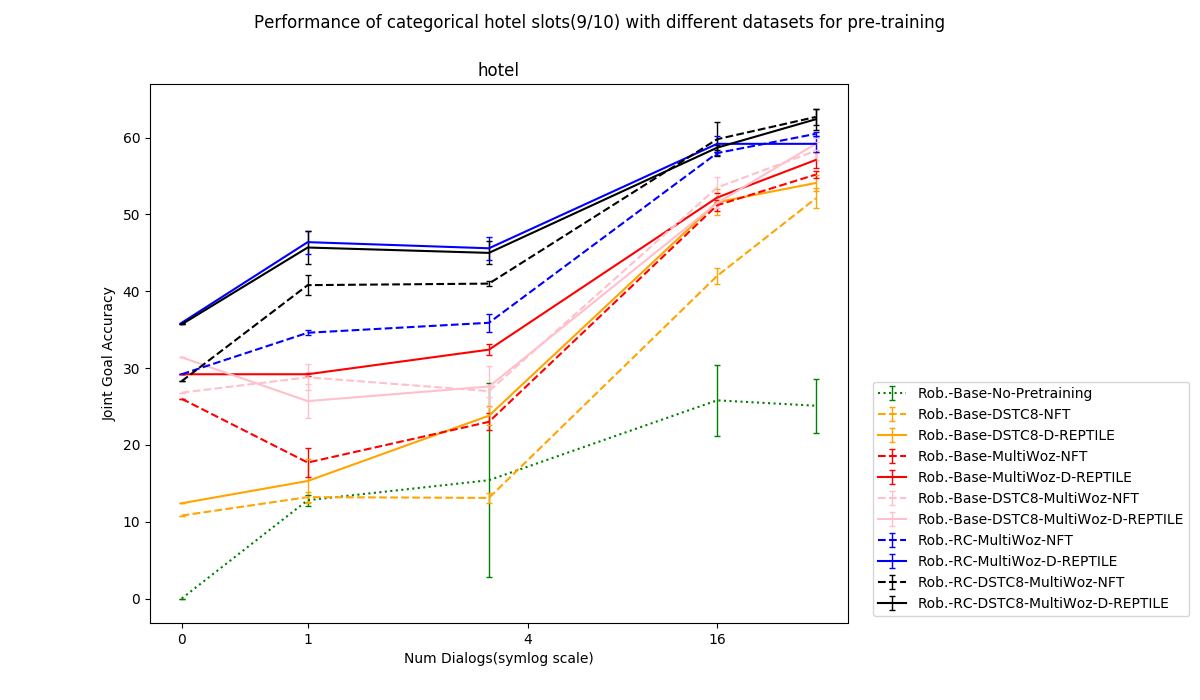}
  \caption{JGA for categorical model for hotel domain with different datasets}
  \label{fig:mix_data}
\end{figure}

\begin{figure}
    \centering
  \includegraphics[width=0.85\linewidth]{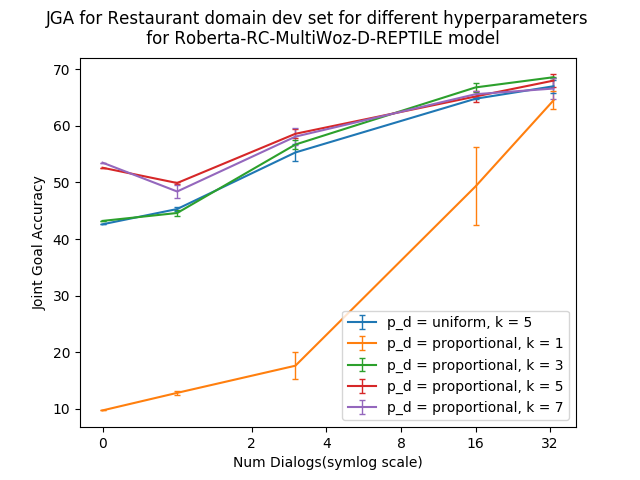}
  \caption{JGA for restaurant domain dev set with different hyper-parameters for the best D-REPTILE model}
  \label{fig:hyper}
\end{figure}

\subsection{Hyper-parameter Search}
We briefly discuss the choice of various hyper-parameters here. We use dev set from restaurant domain for searching for optimum values for different parameters introduced by meta-learning, while the rest are kept same as STARC model \cite{gao2020machine}. In Figure \ref{fig:hyper}, we plot the variation in performance with $k$ and $p_D(.)$. Like any meta-learning algorithm, setting $k$ too small or too large hurts the performance in our case as well (specially $k=1$ where it becomes theoretically similar to NFT \cite{nichol2018first}). Hence, optimum value $k=5$ is used for all our experiments. Also, similar to the conclusion in \citet{dou2019investigating}, we find choosing $p_D(.)$ of any domain as proportional to the size of the training dataset of that domain helpful (blue vs red line). This is attributed to the fact that in case of imbalance in data among different \train domains, the algorithm gets to see all the data from the resource-rich domain as it is chosen more often and hence generalizes better.


\subsection{Adding more \train domains} 
As mentioned in previous section, we observe that benefits of D-REPTILE are much more profound when \target domain is from DSTC8 dataset than when it is from MultiWoz (Figure \ref{fig:cat_dstc8}). Given that DSTC8 has 23 \train domains as compared to 4 in MultiWoz, it is not difficult to see the reason for this boost in performance. In this subsection, we try to answer the question whether MultiWoz \target can also gain from additional domains of DSTC8. Here, for ease of computation, we only experiment with categorical model with \textit{hotel} domain as \target. 
We use both DSTC8 domains and MultiWoz domains (of course excluding hotel domain data during pre-training) and test it on \textit{hotel} data from MultiWoz. These are represented by additional pink and black lines in Figure \ref{fig:mix_data}. We observe that although D-REPTILE helps to improve performance over baseline NFT but adding additional domains does not help the model much overall(solid black line is similar to solid blue line). This shows that in addition to the number of different training tasks, the relatedness of those tasks is also very crucial for meta-learning. The DSTC8 domains which are out-of-sample for MultiWoz \target domain did not prove to be effective. (the small difference between JGA values for 1-dialogue fine-tuning in Figure \ref{fig:mix_data} and categorical model in Figure \ref{fig:cat_non_cat} is due to difference in the choice of the single dialogue from hotel domain used for fine-tuning)


\section{Conclusion}
We conclude our analysis on the merits of meta-learning as compared to naive pre-training for DST problem on a very positive note. Given the practical applicability of very-low data analysis, we provide enough evidence to a developer of an automated conversational system for an unknown domain that irrespective of his/her model and \target domain, D-REPTILE can achieve significant improvement (sometimes almost double) over conventional fine-tuning methods with no additional cost. With detailed ablations, we further provide insights on which slots and  domains will particularly benefit from pre-traning strategies and which of those will require additional data. Being agnostic to underlying model, our proposed algorithm has capability to push state-of-the-art in zero/few-shot DST problem, giving hope for expanding the scope of similar chatbot based systems in new businesses.
\bibliographystyle{acl_natbib}
\bibliography{ref}

\begin{thebibliography}{27}
\expandafter\ifx\csname natexlab\endcsname\relax\def\natexlab#1{#1}\fi

\bibitem[{Bansal et~al.(2019)Bansal, Jha, and McCallum}]{bansal2019learning}
Trapit Bansal, Rishikesh Jha, and Andrew McCallum. 2019.
\newblock Learning to few-shot learn across diverse natural language
  classification tasks.
\newblock \emph{arXiv preprint arXiv:1911.03863}.

\bibitem[{Budzianowski et~al.(2018)Budzianowski, Wen, Tseng, Casanueva, Ultes,
  Ramadan, and Ga{\v{s}}i{\'c}}]{budzianowski2018multiwoz}
Pawe{\l} Budzianowski, Tsung-Hsien Wen, Bo-Hsiang Tseng, Inigo Casanueva,
  Stefan Ultes, Osman Ramadan, and Milica Ga{\v{s}}i{\'c}. 2018.
\newblock Multiwoz-a large-scale multi-domain wizard-of-oz dataset for
  task-oriented dialogue modelling.
\newblock \emph{arXiv preprint arXiv:1810.00278}.

\bibitem[{Campagna et~al.(2020)Campagna, Foryciarz, Moradshahi, and
  Lam}]{campagna2020zero}
Giovanni Campagna, Agata Foryciarz, Mehrad Moradshahi, and Monica~S Lam. 2020.
\newblock Zero-shot transfer learning with synthesized data for multi-domain
  dialogue state tracking.
\newblock \emph{arXiv preprint arXiv:2005.00891}.

\bibitem[{Dai et~al.(2020)Dai, Li, Tang, Li, Sun, and Zhu}]{dai2020learning}
Yinpei Dai, Hangyu Li, Chengguang Tang, Yongbin Li, Jian Sun, and Xiaodan Zhu.
  2020.
\newblock Learning low-resource end-to-end goal-oriented dialog for fast and
  reliable system deployment.
\newblock In \emph{Proceedings of the 58th Annual Meeting of the Association
  for Computational Linguistics}, pages 609--618.

\bibitem[{Dou et~al.(2019)Dou, Yu, and Anastasopoulos}]{dou2019investigating}
Zi-Yi Dou, Keyi Yu, and Antonios Anastasopoulos. 2019.
\newblock Investigating meta-learning algorithms for low-resource natural
  language understanding tasks.
\newblock \emph{arXiv preprint arXiv:1908.10423}.

\bibitem[{Eric et~al.(2019)Eric, Goel, Paul, Kumar, Sethi, Ku, Goyal, Agarwal,
  Gao, and Hakkani-Tur}]{eric2019multiwoz}
Mihail Eric, Rahul Goel, Shachi Paul, Adarsh Kumar, Abhishek Sethi, Peter Ku,
  Anuj~Kumar Goyal, Sanchit Agarwal, Shuyang Gao, and Dilek Hakkani-Tur. 2019.
\newblock Multiwoz 2.1: A consolidated multi-domain dialogue dataset with state
  corrections and state tracking baselines.
\newblock \emph{arXiv preprint arXiv:1907.01669}.

\bibitem[{Finn et~al.(2017)Finn, Abbeel, and Levine}]{finn2017model}
Chelsea Finn, Pieter Abbeel, and Sergey Levine. 2017.
\newblock Model-agnostic meta-learning for fast adaptation of deep networks.
\newblock \emph{arXiv preprint arXiv:1703.03400}.

\bibitem[{Fisch et~al.(2019)Fisch, Talmor, Jia, Seo, Choi, and
  Chen}]{fisch2019mrqa}
Adam Fisch, Alon Talmor, Robin Jia, Minjoon Seo, Eunsol Choi, and Danqi Chen.
  2019.
\newblock Mrqa 2019 shared task: Evaluating generalization in reading
  comprehension.
\newblock \emph{arXiv preprint arXiv:1910.09753}.

\bibitem[{Gao et~al.(2020)Gao, Agarwal, Chung, Jin, and
  Hakkani-Tur}]{gao2020machine}
Shuyang Gao, Sanchit Agarwal, Tagyoung Chung, Di~Jin, and Dilek Hakkani-Tur.
  2020.
\newblock From machine reading comprehension to dialogue state tracking:
  Bridging the gap.
\newblock \emph{arXiv preprint arXiv:2004.05827}.

\bibitem[{Goel et~al.(2019)Goel, Paul, and Hakkani-T{\"u}r}]{goel2019hyst}
Rahul Goel, Shachi Paul, and Dilek Hakkani-T{\"u}r. 2019.
\newblock Hyst: A hybrid approach for flexible and accurate dialogue state
  tracking.
\newblock \emph{arXiv preprint arXiv:1907.00883}.

\bibitem[{Gu et~al.(2018)Gu, Wang, Chen, Cho, and Li}]{gu2018meta}
Jiatao Gu, Yong Wang, Yun Chen, Kyunghyun Cho, and Victor~OK Li. 2018.
\newblock Meta-learning for low-resource neural machine translation.
\newblock \emph{arXiv preprint arXiv:1808.08437}.

\bibitem[{Heck et~al.(2020)Heck, van Niekerk, Lubis, Geishauser, Lin, Moresi,
  and Ga{\v{s}}i{\'c}}]{heck2020trippy}
Michael Heck, Carel van Niekerk, Nurul Lubis, Christian Geishauser, Hsien-Chin
  Lin, Marco Moresi, and Milica Ga{\v{s}}i{\'c}. 2020.
\newblock Trippy: A triple copy strategy for value independent neural dialog
  state tracking.
\newblock \emph{arXiv preprint arXiv:2005.02877}.

\bibitem[{Huang et~al.(2018)Huang, Wang, Singh, Yih, and He}]{huang2018natural}
Po-Sen Huang, Chenglong Wang, Rishabh Singh, Wen-tau Yih, and Xiaodong He.
  2018.
\newblock Natural language to structured query generation via meta-learning.
\newblock \emph{arXiv preprint arXiv:1803.02400}.

\bibitem[{Kingma and Ba(2014)}]{kingma2014adam}
Diederik~P Kingma and Jimmy Ba. 2014.
\newblock Adam: A method for stochastic optimization.
\newblock \emph{arXiv preprint arXiv:1412.6980}.

\bibitem[{Lai et~al.(2017)Lai, Xie, Liu, Yang, and Hovy}]{lai2017race}
Guokun Lai, Qizhe Xie, Hanxiao Liu, Yiming Yang, and Eduard Hovy. 2017.
\newblock Race: Large-scale reading comprehension dataset from examinations.
\newblock \emph{arXiv preprint arXiv:1704.04683}.

\bibitem[{Lee et~al.(2019)Lee, Lee, and Kim}]{lee2019sumbt}
Hwaran Lee, Jinsik Lee, and Tae-Yoon Kim. 2019.
\newblock Sumbt: Slot-utterance matching for universal and scalable belief
  tracking.
\newblock \emph{arXiv preprint arXiv:1907.07421}.

\bibitem[{Liu et~al.(2019)Liu, Ott, Goyal, Du, Joshi, Chen, Levy, Lewis,
  Zettlemoyer, and Stoyanov}]{liu2019roberta}
Yinhan Liu, Myle Ott, Naman Goyal, Jingfei Du, Mandar Joshi, Danqi Chen, Omer
  Levy, Mike Lewis, Luke Zettlemoyer, and Veselin Stoyanov. 2019.
\newblock Roberta: A robustly optimized bert pretraining approach.
\newblock \emph{arXiv preprint arXiv:1907.11692}.

\bibitem[{Mosbach et~al.(2020)Mosbach, Andriushchenko, and
  Klakow}]{mosbach2020stability}
Marius Mosbach, Maksym Andriushchenko, and Dietrich Klakow. 2020.
\newblock On the stability of fine-tuning bert: Misconceptions, explanations,
  and strong baselines.
\newblock \emph{arXiv preprint arXiv:2006.04884}.

\bibitem[{Nichol et~al.(2018)Nichol, Achiam, and Schulman}]{nichol2018first}
Alex Nichol, Joshua Achiam, and John Schulman. 2018.
\newblock On first-order meta-learning algorithms.
\newblock \emph{arXiv preprint arXiv:1803.02999}.

\bibitem[{Qian and Yu(2019)}]{qian2019domain}
Kun Qian and Zhou Yu. 2019.
\newblock Domain adaptive dialog generation via meta learning.
\newblock \emph{arXiv preprint arXiv:1906.03520}.

\bibitem[{Rastogi et~al.(2019)Rastogi, Zang, Sunkara, Gupta, and
  Khaitan}]{rastogi2019towards}
Abhinav Rastogi, Xiaoxue Zang, Srinivas Sunkara, Raghav Gupta, and Pranav
  Khaitan. 2019.
\newblock Towards scalable multi-domain conversational agents: The
  schema-guided dialogue dataset.
\newblock \emph{arXiv preprint arXiv:1909.05855}.

\bibitem[{Sennrich and Zhang(2019)}]{sennrich2019revisiting}
Rico Sennrich and Biao Zhang. 2019.
\newblock Revisiting low-resource neural machine translation: A case study.
\newblock \emph{arXiv preprint arXiv:1905.11901}.

\bibitem[{Sun et~al.(2019)Sun, Yu, Chen, Yu, Choi, and Cardie}]{sun2019dream}
Kai Sun, Dian Yu, Jianshu Chen, Dong Yu, Yejin Choi, and Claire Cardie. 2019.
\newblock Dream: A challenge data set and models for dialogue-based reading
  comprehension.
\newblock \emph{Transactions of the Association for Computational Linguistics},
  7:217--231.

\bibitem[{Vaswani et~al.(2017)Vaswani, Shazeer, Parmar, Uszkoreit, Jones,
  Gomez, Kaiser, and Polosukhin}]{vaswani2017attention}
Ashish Vaswani, Noam Shazeer, Niki Parmar, Jakob Uszkoreit, Llion Jones,
  Aidan~N Gomez, {\L}ukasz Kaiser, and Illia Polosukhin. 2017.
\newblock Attention is all you need.
\newblock In \emph{Advances in neural information processing systems}, pages
  5998--6008.

\bibitem[{Wu et~al.(2019)Wu, Madotto, Hosseini-Asl, Xiong, Socher, and
  Fung}]{wu2019transferable}
Chien-Sheng Wu, Andrea Madotto, Ehsan Hosseini-Asl, Caiming Xiong, Richard
  Socher, and Pascale Fung. 2019.
\newblock Transferable multi-domain state generator for task-oriented dialogue
  systems.
\newblock \emph{arXiv preprint arXiv:1905.08743}.

\bibitem[{Yan et~al.(2020)Yan, Zhang, Jin, and Zhou}]{yan2020multi}
Ming Yan, Hao Zhang, Di~Jin, and Joey~Tianyi Zhou. 2020.
\newblock Multi-source meta transfer for low resource multiple-choice question
  answering.
\newblock In \emph{Proceedings of the 58th Annual Meeting of the Association
  for Computational Linguistics}, pages 7331--7341.

\bibitem[{Zhang et~al.(2019)Zhang, Hashimoto, Wu, Wan, Yu, Socher, and
  Xiong}]{zhang2019find}
Jian-Guo Zhang, Kazuma Hashimoto, Chien-Sheng Wu, Yao Wan, Philip~S Yu, Richard
  Socher, and Caiming Xiong. 2019.
\newblock Find or classify? dual strategy for slot-value predictions on
  multi-domain dialog state tracking.
\newblock \emph{arXiv preprint arXiv:1910.03544}.

\end{thebibliography}

\end{document}